\documentclass[11pt]{article}

\usepackage[preprint]{acl}

\usepackage{times}
\usepackage{latexsym}

\usepackage{tikz}
\usetikzlibrary{calc}
\usepackage{enumitem}

\usepackage{CJKutf8}
\usepackage{multirow}
\usepackage{multicol}
\usepackage{longtable}
\usepackage{array}
\usepackage{booktabs}

\usepackage[T1]{fontenc}

\usepackage[utf8]{inputenc}

\usepackage{microtype}

\usepackage{inconsolata}

\usepackage{graphicx}

\title{Self-Verification is All You Need\\ To Pass The Japanese Bar Examination}

\author{Andrew Shin \\
  Keio University\\
  \texttt{shin@ics.keio.ac.jp} \\}

\begin{document}

\begin{CJK*}{UTF8}{min}

\maketitle
\begin{abstract}
Despite rapid advances in large language models (LLMs), achieving reliable performance on highly professional and structured examinations remains a significant challenge. The Japanese bar examination is a particularly demanding benchmark, requiring not only advanced legal reasoning but also strict adherence to complex answer formats that involve joint evaluation of multiple propositions. While recent studies have reported improvements by decomposing such questions into simpler true--false judgments, these approaches have not been systematically evaluated under the original exam format and scoring scheme, leaving open the question of whether they truly capture exam-level competence. In this paper, we present a self-verification model trained on a newly constructed dataset that faithfully replicates the authentic format and evaluation scale of the exam. Our model is able to exceed the official passing score when evaluated on the actual exam scale, marking the first demonstration, to our knowledge, of an LLM passing the Japanese bar examination without altering its original question structure or scoring rules. We further conduct extensive comparisons with alternative strategies, including multi-agent inference and decomposition-based supervision, and find that these methods fail to achieve comparable performance. Our results highlight the importance of format-faithful supervision and consistency verification, and suggest that carefully designed single-model approaches can outperform more complex systems in high-stakes professional reasoning tasks. Our dataset and codes are publicly available.\footnote{\url{https://github.com/shinandrew/self_verification}}
\end{abstract}

\section{Introduction}

Large language models (LLMs) have demonstrated remarkable capabilities across a wide range of natural language processing tasks, including question answering \cite{Yue2025ASO,Lehmann2024LargeLM}, summarization \cite{Liu2023OnLT}, and even domain-specific reasoning in areas such as mathematics \cite{Shao2024DeepSeekMathPT, Yang2024Qwen25MathTR} and programming \cite{Jiang2024ASO, ElKishky2025CompetitivePW}. Nevertheless, their performance in highly professional domains, particularly law, remains uneven. Legal reasoning often requires precise interpretation of statutes, careful evaluation of multiple interacting conditions, and strict adherence to task-specific output formats, posing challenges that go beyond surface-level linguistic competence.

The Japanese bar examination (司法試験) represents one of the most demanding legal benchmarks in this regard. In addition to its substantive difficulty, its multiple choice exam (短答式) is characterized by a distinctive question format in which examinees must jointly evaluate multiple statements and select correct combinations under rigid answer constraints. Errors in even a single constituent can invalidate an otherwise plausible answer. As we show later in this paper, base LLMs perform poorly under this evaluation regime, highlighting a substantial gap between general language understanding and exam-level legal competence.

Recent work has sought to improve LLM performance on the Japanese bar examination through dataset construction and task reformulation. Notably, the Japanese Bar Exam Question Answering (JBE-QA) dataset \cite{Cao2025JBEQAJB} decomposes complex exam questions into collections of independent true--false judgments, thereby simplifying the learning problem and enabling more stable training. While such decomposition-based approaches have demonstrated promising results on their own benchmarks, they fundamentally alter the structure of the original exam and do not directly address whether models trained in this manner can succeed when confronted with authentic exam questions and scoring criteria.

In this work, rather than modifying the task to suit the model, we develop a consistency-verifying fine-tuning strategy in which the model is trained to generate an answer and subsequently verify its own prediction in the context of the original question. We also design a dataset that preserves the original exam format. This answer-conditioned verification step leverages the model’s strength as an evaluator of candidate solutions and leads to substantial performance gains without increasing model size or relying on external tools. We also investigate more complex inference strategies, including multi-agent architectures, but find that they do not yield improvements under realistic exam conditions.

Through experiments on the Japanese bar examination, we demonstrate that a single fine-tuned model with self-verification trained on a format-faithful dataset can surpass the official passing threshold on the actual exam scale. Specifically, our model obtains the score of 96 from 2024 Japanese bar exam whose passing score is 93. These results suggest that careful alignment between supervision format and evaluation criteria is crucial for advancing LLM performance in high-stakes professional domains.

\section{Related Work}
Numerous benchmarks have been proposed to evaluate LLMs on legal reasoning tasks. In the United States, prior work has shown that GPT-4 achieves passing-level performance on simulated bar examinations \cite{Katz2024GPT4PT}, substantially outperforming earlier models. Other studies have examined legal reasoning across jurisdictions using datasets such as LegalBench \cite{Guha2023LegalBenchAC}, LawBench \cite{Fei2023LawBenchBL}, and national exam-derived corpora in China \cite{Li2024LexEvalAC}, Korea \cite{Kimyeeun2024DevelopingAP}, and Europe \cite{Chlapanis2024LARECHRAN}. These benchmarks cover a wide range of tasks, including multiple-choice question answering, statute interpretation, case outcome prediction, and legal text entailment. While these efforts demonstrate that modern LLMs can perform competitively on legal tasks, many rely on task reformulation, simplified supervision, or indirect evaluation metrics.

Evaluating language models on the Japanese bar examination has recently attracted increasing attention as a challenging benchmark for legal reasoning in Japanese. The examination covers multiple legal domains, including constitutional law, civil law, and criminal law, and is characterized by questions that require joint evaluation of multiple propositions under strict answer-format constraints. Earlier Japanese legal NLP resources, such as the COLIEE shared tasks \cite{Thanh2020JNLPTD,Thanh2021JNLPTD}, primarily focused on civil law and emphasized subtasks like information retrieval or textual entailment, rather than full exam-style question answering. More recent efforts \cite{Nguyen2025NOWJCOLIEE2A} have attempted to construct datasets directly derived from the bar examination, highlighting both the difficulty of the content and the importance of handling the exam’s unique structure. The most notable recent benchmark is the Japanese Bar Exam Question Answering (JBE-QA) dataset \cite{Cao2025JBEQAJB}, which reformulates past bar exam questions into collections of independent true/false statements. Each original question is decomposed into multiple binary judgments, allowing models to be trained and evaluated on simplified supervision signals. Using this formulation, JBE-QA evaluates a wide range of proprietary and open-source LLMs under zero-shot and few-shot settings, and reports substantial performance gains for state-of-the-art models, particularly when chain-of-thought prompting is enabled. While this decomposition strategy stabilizes learning and evaluation, it fundamentally alters the structure of the original exam. As a result, it remains unclear whether models trained under this paradigm can succeed when confronted with intact exam questions that require reasoning over multiple interacting propositions and adherence to the original scoring rules. Our work directly addresses this gap by evaluating models on the authentic exam format and scale.

Recent work has explored multi-agent architectures for legal reasoning \cite{Zhang2025MitigatingMA,Sun2024LawLuoAM}, in which multiple LLM agents collaborate or debate to produce a final answer. Such approaches have been applied to tasks like legal argument generation and multi-step legal analysis, often improving factual coverage or interpretability. However, these systems introduce additional complexity and inference cost, and their effectiveness under strict exam-style evaluation remains underexplored. On the other hand, consistency verification \cite{Patwardhan2024AutomatedCA} and reflection-based methods \cite{Renze2024SelfReflectionIL} have been proposed as general techniques to improve LLM reliability. These approaches encourage a model to evaluate or critique its own output, leveraging the observation that LLMs are often stronger evaluators than generators. 

\begin{table*}[t]
\caption{Comparison between our dataset, and the JBE-QA dataset. While JBE-QA decomposes a single exam question into multiple independent true/false items, our dataset preserves the original joint decision structure. Despite having substantially fewer questions, fine-tuning on our dataset yields significantly higher performance on the actual exam.}
\centering
\scriptsize
\begin{tabular}{p{1.0cm} p{11.0cm} p{0.7cm} p{0.7cm}}
\hline
\textbf{Dataset} & \textbf{Question} & \textbf{Answer} & \textbf{\#Questions} \\
\hline

JBE-QA &
\parbox[t]{11.0cm}{
憲法第３１条の定める法定手続の保障は、直接には刑事手続に関するものであるが、行政手続にも及ぶと解すべき場合があり、その場合には行政処分の相手方に常に事前の告知、弁解、防御の機会を与える必要がある。(Article 31 always requires prior notice and defense opportunity in administrative procedures.)
}
&
False &2,770
 \\\cline{2-3}

 &
\parbox[t]{11.0cm}{
憲法第３５条は、住居、書類及び所持品について、侵入、捜索及び押収を受けることのない権利を規定しているが、この規定の保障対象には、住居、書類及び所持品に準ずる私的領域に侵入されることのない権利が含まれる。(Article 35 protects private domains equivalent to residences, papers, and effects.)
}
&
True & 
 \\\cline{2-3}

 &
\parbox[t]{11.0cm}{
憲法第３８条第１項は、自己が刑事上の責任を問われるおそれのある事項について供述を強要されないことを保障するものであり、氏名の供述も、これによって自己が刑事上の責任を問われるおそれがあることから、原則として保障が及ぶ。(Article 38(1) generally protects refusal to state one’s name.)
}
&
False &
 \\
\hline
\parbox[t]{1.0cm}{Ours\\(identical format as the original exam)} &
\parbox[t]{11.0cm}{
刑事手続上の権利に関する次のアからウまでの各記述について、最高裁判所の判例の趣旨に照らして、それぞれ正しい場合には１を、誤っている場合には２を選びなさい。\\
ア．憲法第３１条の定める法定手続の保障は、直接には刑事手続に関するものであるが、行政手続にも及ぶと解すべき場合があり、その場合には行政処分の相手方に常に事前の告知、弁解、防御の機会を与える必要がある。\\
イ．憲法第３５条は、住居、書類及び所持品について、侵入、捜索及び押収を受けることのない権利を規定しているが、この規定の保障対象には、住居、書類及び所持品に準ずる私的領域に侵入されることのない権利が含まれる。\\
ウ．憲法第３８条第１項は、自己が刑事上の責任を問われるおそれのある事項について供述を強要されないことを保障するものであり、氏名の供述も、これによって自己が刑事上の責任を問われるおそれがあることから、原則として保障が及ぶ。

(Regarding the following statements (A) through (C) concerning rights in criminal procedure, select 1 if the statement is correct in light of Supreme Court precedents, and select 2 if it is incorrect.\\
(A) Article 31 may extend to administrative procedures and always requires prior notice and opportunity to defend.\\
(B) Article 35 protects not only residences, papers, and effects but also equivalent private domains.\\
(C) Article 38(1) protects against compelled self-incrimination, and this protection generally applies to stating one's name.)
}
&
2,1,2 &
460 \\
\hline

\end{tabular}
\label{tab:dataset_comparison}
\end{table*}

\section{Method}
\subsection{Dataset Construction}

We collect actual exam questions spanning 6 years (2019-2024) from the Japanese Ministry of Justice, where we separate 2024 (\textit{Reiwa} 6 or R6) as the test set. We construct a dataset that faithfully replicates the format and evaluation criteria of the Japanese bar examination. Unlike prior work that decomposes questions into independent true/false statements, each instance in our dataset corresponds to a complete exam question, including all constituent statements and the original answer choices. Answers are represented exactly as required in the exam, such as concatenated numeric labels indicating the correctness of each statement or indices corresponding to valid combinations.

Each question is annotated with its subject category (constitutional law(憲法), civil law(民法), or criminal law(刑法)), year of administration, and the points rewarded. This allows evaluation not only in terms of accuracy but also in terms of the official point-based scoring scheme used in the exam. We split the dataset by year, using earlier years (2019-2023, R1-R5) for training and reserving 2024 (R6) exam for evaluation, mirroring realistic exam preparation, where the examinees rely on the past exams for reference. 

Unlike decomposed formulations that reduce each statement to an independent true/false query, correctness in the actual exam depends on the joint evaluation of all statements. For example, an answer may require selecting a single option corresponding to a specific combination of statement truth values, or outputting a concatenated sequence of digits (e.g., ``112'') aligned with multiple statements. An answer is considered incorrect if either the semantic judgment or the required format is violated. Table~\ref{tab:dataset_comparison} contrasts the format of our dataset with that of JBE-QA.

\subsection{Self-Verification}

\begin{figure*}[t]
\centering
\begin{tikzpicture}[
    node distance=4.3cm,
    every node/.style={
        draw,
        rectangle,
        rounded corners,
        align=center,
        font=\scriptsize,
        inner sep=6pt
    },
    arrow/.style={->, thick}
]

\node[text width=3.6cm] (q) {%
\textbf{Exam Question\\ (Original Format)}\\
For each statement, select 1 if it is correct and 2 otherwise.\\
ア.~\ldots \\
イ.~\ldots \\
ウ.~\ldots
};

\node[text width=1.6cm, right of=q] (a1) {%
\textbf{Initial Answer}\\
"112"
};

\node[text width=3.0cm, right of=a1] (verify) {%
\textbf{Self-Verification}\\
``The second answer is incorrect, so modify it to 122.''
};

\node[text width=1.6cm, right of=verify] (a2) {%
\textbf{Final Answer}\\
"122"
};

\draw[arrow] (q) -- (a1);
\draw[arrow] (a1) -- (verify);
\draw[arrow] (verify) -- (a2);

\draw[<->, thick, dashed, looseness=0.5]
(a1.south) to[out=250, in=250]
node[below, font=\scriptsize, draw=none]
{Shared model parameters}
(verify.south);

\end{tikzpicture}
\caption{An overview of our method of self-verification with a shared model under the original exam format.}
\label{fig:overall}
\end{figure*}
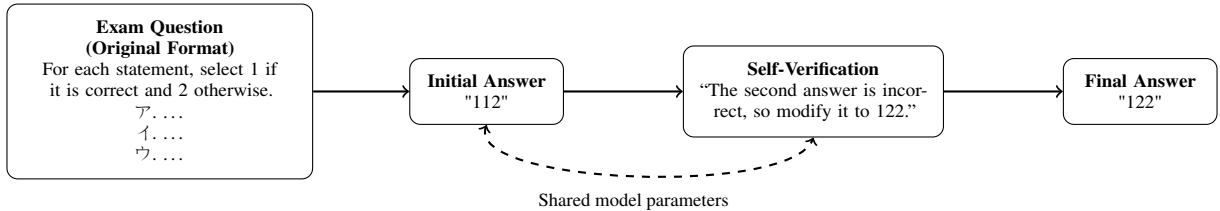

\textbf{Fine-Tuning:} Our approach combines supervised fine-tuning with answer-conditioned self-verification. During training, the model is fine-tuned to generate the correct exam-style answer given the full question, without decomposing the question into simpler subproblems. Given a question $q_i$ consisting of a set of statements $\{{s_i}_1, {s_i}_2, \ldots, {s_i}_n\}$ and a set of valid answer formats defined by the Japanese bar examination, the model is required to produce a single answer $a_i$ that satisfies both semantic correctness and strict formatting constraints, where $a_i$ may contain multiple integers as in the original exam.

\textbf{Self-Verification:} At inference time, we introduce a verification step in which the model re-evaluates its own predicted answer in the context of the original question. Importantly, this verification is performed by the same fine-tuned model, but under a different prompt that induces verification-oriented behavior.

Formally, let $f_\theta(q)$ denote the model’s initial prediction for question $q$. We then define a verification function $g_\theta(q, f_\theta(q))$, which produces a revised answer by assessing the consistency between the question and the initially predicted answer. The final output is given by $\hat{a} = g_\theta(q, f_\theta(q))$. Although $f_\theta$ and $g_\theta$ share the same parameters, they are instantiated with distinct prompts, where one encouraging answer generation and the other encouraging conservative correction. This procedure incurs only a single additional forward pass at inference time, while substantially improving robustness against formatting errors and local reasoning mistakes. Figure~\ref{fig:overall} describes the overall workflow of our approach.

\textbf{Prompt Design:} Table~\ref{tab:prompts} summarizes the prompts used in our system. Answer format instruction prompts have been calibrated to enforce the strict formatting required, which eradicates the necessity for normalization schemes to account for various output formats that frequently occur when bar exam questions are asked. The verification prompt explicitly instructs the model to preserve the original answer unless a clear inconsistency is detected, which we find crucial for preventing unnecessary corrections.

\begin{table*}[t]
\centering
\caption{Prompts used for system role, answer format, and self-verification.}
\scriptsize
\begin{tabular}{p{1.8cm} p{6.5cm} p{6.5cm}}
\hline
\textbf{Purpose} & \textbf{Japanese Prompt} & \textbf{English Translation} \\
\hline
System Role &
あなたは日本の司法試験を受験する受験者である。 &
You are a test taker solving the Japanese bar examination. \\\hline

Answer Format&
【回答形式の厳守】必ず「答えのみ」を出力せよ。理由・説明・記号は一切不要。

1) 選択肢が番号で与えられている場合
   （例：1. アO イO ウO、2. アO イO ウX …）
   → 正しい選択肢の番号のみ出力（例：2）
   
2) 各記述（ア・イ・ウ…）について 1 / 2 を答える問題の場合
   → 数字列のみ出力（例：112）
   
禁止：

- OOX

- アO イO ウX

- ア1 イ1 ウ2

- 説明文 &
【Strict answer format】Output only the answer. Do not include any reasons, explanations, or symbols. 
1) When the choices are given as numbered options (Example: 1. A○ B○ C○, 2. A○ B○ C× …) → Output only the number of the correct option (Example: 2)
2) When each statement (A, B, C, …) requires an answer of 1 or 2
→ Output only the sequence of numbers (Example: 112)
Prohibited:

- OOX

- A○ B○ C×

- A1 B1 C2

- Any explanatory text\\\hline
Verification &
あなたは法律試験の答案を最終確認する役割である。以下の【問題】と【あなたの解答】を照らし合わせ、選択肢番号または数値の形式として最も正しい最終解答を一つだけ出力せよ。

・問題文の条件に照らして明らかに誤っている場合のみ修正すること

・元の解答が正しい場合は、そのまま同じ解答を出力すること

・理由や説明は一切出力せず、最終的な数字のみを出力せよ &
You are responsible for the final review of a law exam answer. Compare the following [Question] and [Your Answer], and output only one final answer in the form of a choice number or numeric value.

・Modify the answer only if it is clearly incorrect based on the question’s conditions.

・If the original answer is correct, output the same answer as is.

・Do not include any reasons or explanations. Output only the final number. \\
\hline
\end{tabular}
\label{tab:prompts}
\end{table*}

\section{Experiments}

\subsection{Experimental Setting}

We evaluate our method on the Japanese bar examination questions from Reiwa 6 (R6), using questions from earlier years (R1--R5) exclusively for fine-tuning. We use GPT-4.1 \cite{OpenAI2025} as our base model, and examine both zero-shot and few-shot setting. For few-shot setting, we chose 5 sample demonstrations from the training set. We also fine-tune separate GPT-4.1 models, with our dataset and with JBE-QA respectively. In order to examine the effect of self-verification, we report the results obtained both with and without self-verification. The same exact prompts were used for all models. For each variant of the models, the experiments were repeated for 3 times.

We report exact-match accuracy as well as the official examination point score, which awards partial credit according to the exam’s grading rules. Partial credit scheme works as following; when 3 or more questions are grouped together with $n$ points, getting one question wrong results in $n-2$ points, whereas getting two or more questions wrong result in 0 point. For example, if 5 questions are grouped together with 4 points, getting 4 questions correct yields 2 points, but getting 3 questions correct results in 0 point, despite the accuracy being over 50\%.

The exam consists of 50 points for constitutional law, 75 points for civil law, and 50 points for criminal law, adding up to 175 maximum points. In case of Reiwa 6 (R6) exam held in 2024, the actual passing score was 93. There is also an additional requirement that at least 40\% of the points should be achieved in each law section.

\subsection{Results \& Analysis}

\begin{table*}[t]
\centering
\caption{Performance comparison on the Japanese bar examination (Reiwa 6). Accuracy denotes exact-match answer accuracy, while scores follow the official exam grading scheme with partial credit. Average subject-wise scores are reported for constitutional law (憲法), civil law (民法), and criminal law (刑法).}
\small
\begin{tabular}{lcccccc}
\hline
\textbf{Model} 
& \textbf{Accuracy} 
& \textbf{Exam Scale (Avg/Min/Max)} 
& \textbf{Const.} 
& \textbf{Civ.} 
& \textbf{Crim.} \\
\hline

Passing Score for Examinees & N/A & 93 (out of 175)& 20 & 30 & 20\\ \hline

Base (Zero-Shot) 
& 0.4036
& 67.0 / 65 / 68
& 8.0
& 32.0 
& 27.0 \\

Base (Few-Shot) 
& 0.3896
& 68.3 / 63 / 71
& 8.0
& 33.3 
& 27.0 \\

Base (Few-Shot) + Self-Verification
& 0.4156
& 76.3 / 76 / 77
& 9.7 
& 36.7 
& \textbf{30.0} \\

Fine-Tuned w/ JBE-QA 
& 0.3766
& 64.0 / 62 / 66
& 8.0 
& 30.0 
& 26.0 \\

Fine-Tuned w/ JBE-QA + Self-Verification 
& 0.4226
& 80.7 / 78 / 82
& 21.0 
& 32.7 
& 27.0 \\

Fine-Tuned w/ Ours 
& 0.4675 
& 92.3 / 91 / 93
& 20.3 
& 42.0 
& \textbf{30.0} \\

Fine-Tuned w/ Ours + Self-Verification 
& \textbf{0.4935} 
& \textbf{94.7 / 94 / 96} 
& \textbf{22.3} 
& \textbf{42.3} 
& \textbf{30.0} \\\hline

Multi-Agent (same model for all agents)
& 0.4026 
& 75.7 / 74 / 79
& 19.3 
& 30.7 
& 25.7 \\

Multi-Agent (separately fine-tuned models)
& 0.3969
& 71.0 / 66 / 77
& 12.7
& 34.7
& 25.6\\
\hline
\end{tabular}
\label{tab:main_results}
\end{table*}

Table~\ref{tab:main_results} summarizes the results from the models examined. 

\textbf{Base models:} Base model with zero-shot setting performs poorly, clearly suggesting that the legal knowledge on which the model has been pre-trained is insufficient for a reasonable performance on exam-level tasks. Few-shot setting hardly boosts the performance over zero-shot. While it is expected that providing a few samples would not significantly improve the legal knowledge, it also does not particularly seem to have helped in guiding exam-specific format. A clear performance boost is made by performing self-verification on base model. In fact, as we shall see, self-verification invariably boosts the performance regardless of model choice, demonstrating it is an efficient model-agnostic technique.

\textbf{JBE-QA:} Despite being trained on a substantially larger dataset, the model fine-tuned on JBE-QA exhibits markedly poor performance on the actual Japanese bar examination, suggesting that improving the model's legal knowledge does not automatically translate to performance boost in more complex tasks. While their decomposition strategy simplifies learning by isolating local factual judgments, it removes the requirement to reason over joint constraints among statements. The true/false reformulation introduces an implicit shift in task distribution, as the model is optimized for binary classification rather than constrained selection under combinatorial rules. As such, fine-tuning on decomposed propositions encourages a form of segmented knowledge representation that lacks mechanisms for re-composition at inference time.



These findings suggest that while proposition-level supervision may improve performance on simplified benchmarks, it does not necessarily transfer to evaluation settings that require holistic reasoning. For high-stakes professional examinations, preserving the native question format during training appears critical for enabling models to align local legal knowledge with global decision consistency.

As with the base models, the model fine-tuned with JBE-QA shows a significant performance boost with self-verification. This again reinforces that self-verification is an effective technique regardless of the model.

\textbf{Ours:}
Fine-tuning with our dataset, with or without self-verification, clearly outperforms other approaches. Notably, fine-tuning with our dataset alone without self-verification already obtains scores around the passing score. Note that its performance gain cannot be attributed to format memorization or answer pattern learning, as the combinatorial space of such answers (e.g. "11221”) makes correct prediction by memorizing answer frequencies or guessing implausible. Moreover, the training data does not provide decomposed supervision at the proposition level; the model must internally reason over each constituent statement to produce a globally consistent output. This suggests that exposure to the authentic multi-proposition format during fine-tuning induces an ability to jointly assess multiple legal conditions within a single reasoning context. This is further supported by the fact that models trained on decomposed dataset fail to recover comparable performance when evaluated under the original exam structure, despite having access to substantially more training instances.

As with other model variants, we observe a further and consistent performance improvement by introducing self-verification at inference time. While the fine-tuned model already produces strong initial answers, the verification step allows the model to reassess the internal consistency of its own prediction against the original question, particularly for answers involving multiple propositions. This process is effective at correcting local inconsistencies, such as a single misjudged statement within an otherwise correct composite answer, which directly translates into higher exam scores under the official grading scheme. Importantly, self-verification does not introduce external supervision or additional training data, but instead leverages the model’s existing legal knowledge in a second-pass reasoning phase. The resulting improvement suggests that a non-trivial portion of errors made by strong fine-tuned models arise not from lack of legal knowledge, but from failures in global consistency, which self-verification is well suited to mitigate.

Another plausible explanation is that both format-specific fine-tuning and self-verification act as catalysts that elicit latent knowledge already present in the model. This may explain why performance improves substantially despite the relatively small size of our training data, which by itself is unlikely to contain all the legal knowledge required to pass the examination. Self-verification further amplifies this effect by encouraging the model to reassess and consolidate its own predictions, as evidenced by the consistent, model-agnostic performance gains observed when verification is applied.

In short, format-specific fine-tuning teaches the model how to exploit internal knowledge that would otherwise remain dormant, and self-verification further strengthens this elicitation process by promoting global consistency across multiple propositions. 

Table~\ref{tab:qualitative_examples_compact} shows qualitative examples with the outputs from each model, along with the points awarded to each output. Our model correctly answers both the single answer format and the composite answer format, which other models struggle to address. Improvements with self-verification can be seen in other models as well, where they receive partial credit. Note that there are instances where other models often produce incompatible outputs, such as the additional number of answers. Also, note that in many cases, 0 points are awarded even when the accuracy for the questions is over 50\%. This reinforces the point that performing well on decomposed propositions does not automatically translate to equivalent performance on actual exam format and scale, and that it requires composite reasoning to demonstrate genuine success, rather than to simply claim competence based on simplified benchmarks.

\begin{table*}[t]
\centering
\scriptsize
\setlength{\tabcolsep}{3pt}
\caption{Qualitative examples under the authentic Japanese bar examination format. For models, +V indicates that self-verification is performed. For each output, points awarded are displayed in parenthesis. Bold outputs indicate correct outputs and maximum points. Outputs that do not match the number of digits are actual mistakes by the models.}
\label{tab:qualitative_examples_compact}

\begin{tabular}{p{0.08\textwidth} p{0.88\textwidth}}
\toprule

\textbf{Question 1} &
憲法第２２条に関する次のアからウまでの各記述について、それぞれ正しい場合には１を、誤っている場合には２を選びなさい。
(\emph{For each of the following statements (A)–(C) concerning Article 22 of the Constitution, select 1 if correct and 2 if incorrect.}) \\

& ア．判例は、日本に適法に在留する外国人には、憲法上、その在留期間内において外国へ一時旅行する自由が保障されているものと解している。
(\emph{Precedent holds that foreign nationals lawfully residing in Japan are constitutionally guaranteed the freedom to temporarily travel abroad during their period of stay.}) \\

& イ．居住・移転の自由は、複合的な性格を有する人権と解されており、広く知的な接触の機会を得るために不可欠であることから、精神的自由の要素も併せ持っている。
(\emph{Freedom of residence and movement is understood as a right with a composite character, and because it is indispensable for obtaining broad opportunities for intellectual contact, it also possesses aspects of spiritual freedom.}) \\

& ウ．判例は、市営住宅の入居者が暴力団員であることが判明したときには当該住宅の明渡しを請求することができるとする条例の規定による居住の制限は、公共の福祉による必要かつ合理的なものであるから、この規定は憲法第２２条第１項に違反しないと解している。
(\emph{Precedent holds that an ordinance allowing eviction from public housing when a resident is found to be a gang member constitutes a necessary and reasonable restriction for the public welfare, and thus does not violate Article 22(1) of the Constitution.}) \\

\midrule

\textbf{Model} &
\begin{tabular}{@{}*{10}{>{\centering\arraybackslash}p{0.077\textwidth}}@{}}
Exam(GT) &
Base(ZS) &
Base(FS) &
Base+V &
JBE-QA &
JBE-QA+V &
Ours &
Ours+V &
MA(Same) &
MA(Sep)
\end{tabular} \\\hline

\textbf{Output (Pts)} &
\begin{tabular}{@{}*{10}{>{\centering\arraybackslash}p{0.077\textwidth}}@{}}
\textbf{211(3)} &
121(0) &
121(0) &
221(1) &
1112(0) &
121(1) &
\textbf{211(3)} &
\textbf{211(3)} &
122(0) &
112(0)
\end{tabular} \\
\midrule\midrule

\textbf{Question 2} &
相続人に関する次のアからオまでの各記述のうち、判例の趣旨に照らし誤っているものを組み合わせたものは、後記１から５までのうちどれか。
(\emph{Which of the following combinations consists of statements that are incorrect in light of Supreme Court precedent?}) \\

& ア．被相続人の内縁の配偶者は、相続人となる。
(\emph{A de facto spouse of the decedent becomes an heir.}) \\

& イ．被相続人が妻の懐胎中に死亡したときは、その後に出生した子は、相続人となる。
(\emph{If the decedent dies while his wife is pregnant, the child subsequently born becomes an heir.}) \\

& ウ．被相続人Ａと子Ｂが死亡し、その前後関係が不明な場合、Ｂの子Ｃは代襲相続する。
(\emph{If both the decedent A and child B die and the order of death is unknown, B’s child C succeeds by representation.}) \\

& エ．子Ｂが相続放棄した場合、Ｂの子Ｃは代襲相続する。
(\emph{If child B renounces inheritance, B’s child C succeeds by representation.}) \\

& オ．遺言書を破棄しても不当目的がなければ相続欠格に当たらない。
(\emph{An heir who destroys a will without unjust intent does not become disqualified from inheritance.}) \\

\midrule

\textbf{Model} &
\begin{tabular}{@{}*{10}{>{\centering\arraybackslash}p{0.077\textwidth}}@{}}
Exam(GT) &
Base(ZS) &
Base(FS) &
Base+V &
JBE-QA &
JBE-QA+V &
Ours &
Ours+V &
MA(Same) &
MA(Sep)
\end{tabular} \\\hline

\textbf{Output (Pts)} &
\begin{tabular}{@{}*{10}{>{\centering\arraybackslash}p{0.077\textwidth}}@{}}
\textbf{2(2)} &
3(0) &
3(0) &
\textbf{2(2)} &
3(0) &
3(0) &
\textbf{2(2)} &
\textbf{2(2)} &
4(0) &
1(0) \\
\end{tabular}\\
\midrule\midrule

\textbf{Question 3} &
次のアからオまでの各記述を判例の立場に従って検討し、正しい場合には１を、誤っている場合には２を選びなさい。(Examine each of the following statements A–E according to judicial precedent; choose 1 if correct and 2 if incorrect.)\\

&ア．甲は、宝くじの当せん金を得るため、外れた宝くじに印字された番号を当せん番号に改ざんした。 この場合、甲に有印私文書変造罪が成立する。  
(For the purpose of obtaining lottery winnings, X altered the number printed on a losing lottery ticket to match the winning number. In this case, the crime of alteration of a private document with a seal is established.)\\

&イ．甲は、事情を知らない乙に対し、偽造通貨を真正な通貨のように装って代金として交付し、乙から商品を購入した。  
この場合、甲に詐欺罪及び偽造通貨行使罪が成立し、両罪は観念的競合となる。  
(X handed counterfeit currency to Y, who was unaware of the falsity, as if it were genuine currency, and purchased goods from Y. In this case, fraud and uttering counterfeit currency are established, and the two crimes are in conceptual concurrence.)\\

&ウ．甲は、乙から、乙の代わりにＡ大学の入学試験を受けてほしいと頼まれ、これを引き受け、乙に成り済まして同入学試験を受け、氏名欄に乙の氏名を記載し、乙名義で答案を作成した。  
この場合、甲に有印私文書偽造罪が成立する。  
(X agreed to take an entrance examination for University A on behalf of Y, impersonated Y, wrote Y’s name, and prepared an answer sheet under Y’s name. In this case, the crime of forgery of a private document with a seal is established.)\\

&エ．甲は、行使の目的で、他人が振り出した額面１００万円の小切手の金額欄に「０」を加え、額面１０００万円の小切手に改ざんした。  
この場合、甲に有価証券偽造罪が成立する。  
(For the purpose of use, X added a zero to the amount field of a 1-million-yen check issued by another person, altering it into a 10-million-yen check. In this case, the crime of forgery of a valuable security is established.)\\

&オ．甲は、乙から金銭の借入れとして１万円札１０枚を受け取った際、それらの中に偽造の１万円札が含まれていることに気付かず、その後、偽造の１万円札の存在に気付いたが、行使の目的でそのまま保持した。  
この場合、甲に偽造通貨収得罪は成立しない。  
(X received ten 10,000-yen bills as a loan from Y without noticing that one was counterfeit; later noticing the counterfeit bill, X continued to possess it for the purpose of use. In this case, the crime of acquisition of counterfeit currency is not established.)
\\
\hline
\textbf{Model} &
\begin{tabular}{@{}*{10}{>{\centering\arraybackslash}p{0.077\textwidth}}@{}}
Exam(GT) &
Base(ZS) &
Base(FS) &
Base+V &
JBE-QA &
JBE-QA+V &
Ours &
Ours+V &
MA(Same) &
MA(Sep)
\end{tabular}
\\
\hline
\textbf{Output (Pts)} &
\begin{tabular}{@{}*{10}{>{\centering\arraybackslash}p{0.077\textwidth}}@{}}
\textbf{22121(4)} &
21222(0) &
21222(0) &
21122(0) &
21212(0) &
21211(0) &
21121(2) &
\textbf{22121(4)} &
21222(0) &
21212(0)
\end{tabular}
\\
\bottomrule
\end{tabular}
\end{table*}

\section{Multi-Agent Reasoning}
Multi-agent reasoning has demonstrated success in a number of complex tasks \cite{Zhang2025AgenticCE}, and has also been attempted in legal domain \cite{Zhang2025MitigatingMA,Sun2024LawLuoAM}. To evaluate whether explicit decomposition into interacting agents improves performance on the Japanese bar examination, we implemented a multi-agent pipeline in which distinct agents are responsible for retrieval, verification, knowledge abstraction, and final answering. Unlike prior work that assumes homogeneous agents or informal collaboration, our implementation assigns clearly separated functional roles and evaluates both shared-model and independently fine-tuned agent configurations under the authentic exam grading scheme.

\begin{table*}[h]
\centering
\caption{Role-specific prompts used in the multi-agent pipeline.}
\small
\begin{tabular}{p{2.8cm} p{9.5cm}}
\hline
\textbf{Agent} & \textbf{Prompt (English translation in parentheses)} \\
\hline
Retriever & 以下の問題に関連すると考えられる過去問とその回答を選択せよ。
選択の基準は、扱われている法分野、論点、条文、または判例の種類が共通しているかどうかである。最大で数問まで選んでよい。（Select past exam questions and answers that you consider relevant to the following problem. Relevance should be judged based on shared legal domain, issues, statutes, or types of precedents. You may select up to a few questions.）\\\hline

Verifier &
以下の問題に対して参考になる過去問と回答のみを選別してください。 
（Select only the past exam questions and answers that are relevant to the following problem.） \\\hline

Extractor &
以下の問題と正解から、将来の類似問題に使える一般化可能な法的知識を抽出せよ。  
（Extract generalizable legal knowledge from the following question and answer that can be reused for future similar problems.） \\\hline

Reasoner &
以下は関連する法的知識である。上記を踏まえて、次の問題に答えよ。  
（Below is relevant legal knowledge. Based on it, answer the following question.） \\

\hline
\end{tabular}
\label{tab:multi_prompts}
\end{table*}

\subsection{Multi-Agent Architecture}

The pipeline consists of four sequential agents, loosely inspired by the architecture proposed by \cite{Zhang2025AgenticCE}; 
\begin{itemize}
    \item \textit{Retriever Agent:}
Given a test question $q$, the retriever agent selects a set of candidate past exam questions and answers from the training set (R1--R5) that it finds relevant to the question $q$. 
\item \textit{Verifier Agent:} The verifier agent receives the test question and the retrieved candidates from the past exams, and filters them to retain only those deemed relevant. Its role is to discard superficially similar but legally irrelevant questions, thereby reducing noise before knowledge abstraction. 
\item \textit{Knowledge Extraction Agent:} For each verified past question, the extraction agent abstracts generalizable legal principles from the question/answer pair. The agent is instructed to output only reusable criteria, conditions, or patterns in bullet-point form. 
\item \textit{Final Reasoning Agent:} The reasoning agent receives the original test question together with the aggregated extracted knowledge and produces the final answer in the strict exam-required format. This agent is solely responsible for joint evaluation of all propositions and adherence to formatting constraints.
\end{itemize}

We evaluated two configurations of this architecture. In the shared-model setting, all four agents were instantiated from the same model fine-tuned with our dataset as in the previous experiment, isolating the effect of role separation and interaction. In the independently fine-tuned setting, each agent was fine-tuned separately on the same training data but with role-specific prompts, with the goal of increasing functional specialization and diversity. Prompts for each agent are shown in Table~\ref{tab:multi_prompts}. Answer format instruction is identical as the previous experiment.

\subsection{Results \& Analysis}

As shown in Table~\ref{tab:main_results}, both configurations performed substantially worse than a single fine-tuned model. The shared-model multi-agent system achieved an average score of 75.7 points, while the independently fine-tuned multi-agent system further degraded performance to 71.0 points, both of which are substantially below the passing score. These results fall far below the single-model approach, despite significantly increased inference complexity.

Our results highlight that multi-agent systems do not automatically outperform strong single-model baselines. On the contrary, in tightly constrained tasks such as the Japanese bar examination, distributing reasoning across agents can be harmful, as errors introduced by individual agents tend to propagate and compound throughout the pipeline. In particular, abstraction and verification stages may introduce subtle inconsistencies that the final reasoning agent must reconcile under strict formatting and joint-consistency constraints.

We further find that increasing agent diversity through independent fine-tuning does not improve performance and instead exacerbates coordination failures. While independently trained agents exhibit greater surface-level variation, they lack a shared representational space that would allow their outputs to be reliably integrated. In contrast, shared representations appear to be crucial for effective agentic behavior in multi-agent setting, especially in settings where success depends on maintaining global consistency across multiple interdependent propositions.

\section{Conclusion \& Future Work}

In this paper, we presented the first large language model system to achieve a passing score on the Japanese bar examination when evaluated under its original question format and official grading scale. Our results demonstrate that preserving the exam’s multi-proposition structure during dataset construction and fine-tuning is essential. A single fine-tuned model, augmented with lightweight self-verification, is sufficient to meet the passing threshold without resorting to question decomposition or external supervision, despite a relatively small size of training data. 

In contrast, approaches based on decomposed propositions or multi-agent deliberation fail to achieve comparable performance when assessed under realistic exam conditions, despite increased architectural complexity. These findings indicate that success on the Japanese bar examination depends less on distributing reasoning across components than on maintaining global consistency over tightly coupled propositions. The additional gains from self-verification further underscore that many remaining errors arise from coherence failures rather than missing legal knowledge. Overall, both format-specific fine-tuning and self-verification act as effective catalysts to extract the latent knowledge present in the model that would not be exploited otherwise.

As future work, we aim to extend this approach to the free-response (論文式) portion of the exam, which requires structured legal argumentation rather than discrete answer selection. More broadly, our results highlight the importance of evaluating legal reasoning systems under authentic task formats and caution against drawing conclusions from benchmarks that substantially simplify the structure and evaluation criteria.

\section*{Limitations}

This work focuses exclusively on the multiple-choice (短答式) portion of the Japanese bar examination and does not address the free-response (論文式) component, which requires structured legal argumentation, citation control, and extended reasoning. As such, our results should not be interpreted as demonstrating comprehensive legal reasoning ability or qualification-level competence. In addition, although our dataset faithfully preserves the original exam format, its size is relatively small compared to large-scale legal benchmarks, and performance gains rely on the presence of substantial prior knowledge in the underlying base model. Our approach therefore assumes access to a strong pretrained language model and may not generalize to weaker or domain-mismatched models. Finally, while self-verification improves robustness under the exam’s grading scheme, it does not guarantee correctness in cases where the model’s internal legal knowledge is itself incorrect or outdated.

\section*{Ethical Statement}
This study does not involve human subjects, personal data, or sensitive private information. All exam questions used in our dataset are publicly available past questions from the Japanese bar examination, used solely for research and evaluation purposes. Our results should not be construed as endorsing the use of language models as a substitute for formal legal education, professional training, or legal advice. In particular, passing or near-passing performance on an exam-style benchmark does not imply real-world legal competence or ethical judgment. We emphasize that any deployment of such models in legal contexts must be accompanied by appropriate human oversight and clear disclosure of limitations. We caution against over-interpreting benchmark success without evaluation under authentic task formats and grading schemes, a concern that our work explicitly aims to highlight rather than exacerbate.

\bibliography{custom}

@inproceedings{Cao2025JBEQAJB,
  title={JBE-QA: Japanese Bar Exam QA Dataset for Assessing Legal Domain Knowledge},
  author={Zhihan Cao and Fumihito Nishino and Hiroaki Yamada and Nguyen Ha Thanh and Yusuke Miyao and Ken Satoh},
  year={2025},
  url={https://api.semanticscholar.org/CorpusID:283438179}
}

@misc{OpenAI2025,
  author = {{OpenAI}},
  howpublished = {\url{https://openai.com/index/gpt-4-1/}},
  title = {Introducing GPT-4.1 in the API},
  year = {2025}
}

@article{Katz2024GPT4PT,
  title={GPT-4 passes the bar exam},
  author={Daniel Martin Katz and Michael James Bommarito and Shang Gao and Pablo Arredondo},
  journal={Philosophical transactions. Series A, Mathematical, physical, and engineering sciences},
  year={2024},
  volume={382},
  url={https://api.semanticscholar.org/CorpusID:257572753}
}

@article{Yue2025ASO,
  title={A Survey of Large Language Model Agents for Question Answering},
  author={Murong Yue},
  journal={ArXiv},
  year={2025},
  volume={abs/2503.19213},
  url={https://api.semanticscholar.org/CorpusID:277313791}
}

@inproceedings{Lehmann2024LargeLM,
  title={Large Language Models for Scientific Question Answering: An Extensive Analysis of the SciQA Benchmark},
  author={Jens Lehmann and Antonello Meloni and Enrico Motta and Francesco Osborne and Diego Reforgiato Recupero and Angelo Salatino and Sahar Vahdati and TU ScaDS.AI- and Dresden and De},
  booktitle={Extended Semantic Web Conference},
  year={2024},
  url={https://api.semanticscholar.org/CorpusID:269767509}
}

@inproceedings{Liu2023OnLT,
  title={On Learning to Summarize with Large Language Models as References},
  author={Yixin Liu and Alexander R. Fabbri and Pengfei Liu and Dragomir R. Radev and Arman Cohan},
  booktitle={North American Chapter of the Association for Computational Linguistics},
  year={2023},
  url={https://api.semanticscholar.org/CorpusID:258841126}
}

@article{ElKishky2025CompetitivePW,
  title={Competitive Programming with Large Reasoning Models},
  author={Ahmed El-Kishky and Alexander Wei and Andre Saraiva and Borys Minaev and Daniel Selsam and David Dohan and Francis Song and Hunter Lightman and Ignasi Clavera and Jakub W. Pachocki and Jerry Tworek and Lorenz Kuhn and Lukasz Kaiser and Mark Chen and Max Schwarzer and Mostafa Rohaninejad and Nat McAleese and o3 contributors and Oleg Murk and Rhythm Garg and Rui Shu and Szymon Sidor and Vineet Kosaraju and Wenda Zhou},
  journal={ArXiv},
  year={2025},
  volume={abs/2502.06807},
  url={https://api.semanticscholar.org/CorpusID:276258630}
}

@article{Shao2024DeepSeekMathPT,
  title={DeepSeekMath: Pushing the Limits of Mathematical Reasoning in Open Language Models},
  author={Zhihong Shao and Peiyi Wang and Qihao Zhu and Runxin Xu and Jun-Mei Song and Mingchuan Zhang and Y. K. Li and Yu Wu and Daya Guo},
  journal={ArXiv},
  year={2024},
  volume={abs/2402.03300},
  url={https://api.semanticscholar.org/CorpusID:267412607}
}

@article{Yang2024Qwen25MathTR,
  title={Qwen2.5-Math Technical Report: Toward Mathematical Expert Model via Self-Improvement},
  author={An Yang and Beichen Zhang and Binyuan Hui and Bofei Gao and Bowen Yu and Chengpeng Li and Dayiheng Liu and Jianhong Tu and Jingren Zhou and Junyang Lin and Keming Lu and Mingfeng Xue and Runji Lin and Tianyu Liu and Xingzhang Ren and Zhenru Zhang},
  journal={ArXiv},
  year={2024},
  volume={abs/2409.12122},
  url={https://api.semanticscholar.org/CorpusID:272707652}
}

@article{Jiang2024ASO,
  title={A Survey on Large Language Models for Code Generation},
  author={Juyong Jiang and Fan Wang and Jiasi Shen and Sungju Kim and Sunghun Kim},
  journal={ACM Transactions on Software Engineering and Methodology},
  year={2024},
  url={https://api.semanticscholar.org/CorpusID:270214176}
}

@article{Guha2023LegalBenchAC,
  title={LegalBench: A Collaboratively Built Benchmark for Measuring Legal Reasoning in Large Language Models},
  author={Neel Guha and Julian Nyarko and Daniel E. Ho and Christopher R{\'e} and Adam Chilton and Aditya Narayana and Alex Chohlas-Wood and Austin M. K. Peters and Brandon Waldon and Daniel N. Rockmore and Diego A. Zambrano and Dmitry Talisman and Enam Hoque and Faiz Surani and Frank Fagan and Galit Sarfaty and Gregory M. Dickinson and Haggai Porat and Jason Hegland and Jessica Wu and Joe Nudell and Joel Niklaus and John J. Nay and Jonathan H. Choi and Kevin Patrick Tobia and Margaret Hagan and Megan Ma and Michael A. Livermore and Nikon Rasumov-Rahe and Nils Holzenberger and Noam Kolt and Peter Henderson and Sean Rehaag and Sharad Goel and Shangsheng Gao and Spencer Williams and Sunny G. Gandhi and Tomer Zur and Varun J. Iyer and Zehua Li},
  journal={ArXiv},
  year={2023},
  volume={abs/2308.11462},
  url={https://api.semanticscholar.org/CorpusID:261064672}
}

@article{Fei2023LawBenchBL,
  title={LawBench: Benchmarking Legal Knowledge of Large Language Models},
  author={Zhiwei Fei and Xiaoyu Shen and D. Zhu and Fengzhe Zhou and Zhuo Han and Songyang Zhang and Kai Chen and Zongwen Shen and Jidong Ge},
  journal={ArXiv},
  year={2023},
  volume={abs/2309.16289},
  url={https://api.semanticscholar.org/CorpusID:263134950}
}

@article{Li2024LexEvalAC,
  title={LexEval: A Comprehensive Chinese Legal Benchmark for Evaluating Large Language Models},
  author={Haitao Li and You Chen and Qingyao Ai and Yueyue Wu and Ruizhe Zhang and Yiqun Liu},
  journal={ArXiv},
  year={2024},
  volume={abs/2409.20288},
  url={https://api.semanticscholar.org/CorpusID:272987186}
}

@inproceedings{Kimyeeun2024DevelopingAP,
  title={Developing a Pragmatic Benchmark for Assessing Korean Legal Language Understanding in Large Language Models},
  author={Kimyeeun Kimyeeun and Choi Youngrok and Eunkyung Choi and Jinhwan Choi and Hai Jin Park and Wonseok Hwang},
  booktitle={Conference on Empirical Methods in Natural Language Processing},
  year={2024},
  url={https://api.semanticscholar.org/CorpusID:273323722}
}

@article{Chlapanis2024LARECHRAN,
  title={LAR-ECHR: A New Legal Argument Reasoning Task and Dataset for Cases of the European Court of Human Rights},
  author={Odysseas S. Chlapanis and Dimitrios Galanis and Ion Androutsopoulos},
  journal={ArXiv},
  year={2024},
  volume={abs/2410.13352},
  url={https://api.semanticscholar.org/CorpusID:273404312}
}

@article{Nguyen2025NOWJCOLIEE2A,
  title={NOWJ@COLIEE 2025: A Multi-stage Framework Integrating Embedding Models and Large Language Models for Legal Retrieval and Entailment},
  author={Hoang-Trung Nguyen and Tan-Minh Nguyen and Xuan-Bach Le and Tuan-Kiet Le and Khanh-Huyen Nguyen and Ha Thanh Nguyen and Thi-Hai-Yen Vuong and Le-Minh Nguyen},
  journal={ArXiv},
  year={2025},
  volume={abs/2509.08025},
  url={https://api.semanticscholar.org/CorpusID:281244268}
}

@article{Zhang2025MitigatingMA,
  title={Mitigating Manipulation and Enhancing Persuasion: A Reflective Multi-Agent Approach for Legal Argument Generation},
  author={Li Zhang and Kevin Ashley},
  journal={ArXiv},
  year={2025},
  volume={abs/2506.02992},
  url={https://api.semanticscholar.org/CorpusID:279118723}
}

@inproceedings{Sun2024LawLuoAM,
  title={LawLuo: A Multi-Agent Collaborative Framework for Multi-Round Chinese Legal Consultation},
  author={Jingyun Sun and Chengxiao Dai and Zhongze Luo and Yangbo Chang and Yang Li},
  year={2024},
  url={https://api.semanticscholar.org/CorpusID:271334345}
}

@article{Patwardhan2024AutomatedCA,
  title={Automated Consistency Analysis of LLMs},
  author={Aditya Patwardhan and Vivek Vaidya and Ashish Kundu},
  journal={2024 IEEE 6th International Conference on Trust, Privacy and Security in Intelligent Systems, and Applications (TPS-ISA)},
  year={2024},
  pages={118-127},
  url={https://api.semanticscholar.org/CorpusID:275583546}
}

@article{Renze2024SelfReflectionIL,
  title={Self-Reflection in Large Language Model Agents: Effects on Problem-Solving Performance},
  author={Matthew Renze and Erhan Guven},
  journal={2024 2nd International Conference on Foundation and Large Language Models (FLLM)},
  year={2024},
  pages={516-525},
  url={https://api.semanticscholar.org/CorpusID:275956778}
}

@article{Zhang2025AgenticCE,
  title={Agentic Context Engineering: Evolving Contexts for Self-Improving Language Models},
  author={Qizheng Zhang and Changran Hu and Shubhangi Upasani and Boyuan Ma and Fenglu Hong and Vamsidhar Reddy Kamanuru and Jay Rainton and Chen Wu and Mengmeng Ji and Hanchen Li and Urmish Thakker and James Zou and Kunle Olukotun},
  journal={ArXiv},
  year={2025},
  volume={abs/2510.04618},
  url={https://api.semanticscholar.org/CorpusID:281842432}
}

@article{Thanh2021JNLPTD,
  title={JNLP Team: Deep Learning Approaches for Legal Processing Tasks in COLIEE 2021},
  author={Nguyen Ha Thanh and Phuong Minh Nguyen and Thi-Hai-Yen Vuong and Quan Minh Bui and Chau Nguyen and Binh Dang and Vu Tran and Minh Le Nguyen and Ken Satoh},
  journal={ArXiv},
  year={2021},
  volume={abs/2106.13405},
  url={https://api.semanticscholar.org/CorpusID:235651986}
}

@article{Thanh2020JNLPTD,
  title={JNLP Team: Deep Learning for Legal Processing in COLIEE 2020},
  author={Nguyen Ha Thanh and Hai-Yen Thi Vuong and Phuong Minh Nguyen and Binh Dang and Quan Minh Bui and Sinh Trong Vu and Chau Nguyen and Vu Tran and Ken Satoh and Minh Le Nguyen},
  journal={ArXiv},
  year={2020},
  volume={abs/2011.08071},
  url={https://api.semanticscholar.org/CorpusID:226965686}
}



\end{CJK*}

\end{document}